\documentclass[10pt, a4paper]{article}

\usepackage{amsmath}
\usepackage[final]{lrec2026} 
\usepackage{booktabs}
\usepackage{multirow}

\usepackage{float}

\title{From Articles to Premises: Building PrimeFacts, an Extraction Methodology and Resource for Fact-Checking Evidence}

\name{{\bfseries \large Premtim Sahitaj$^{1,2}$, Jawan Kolanowski$^3$, Ariana Sahitaj$^{1,2}$} \\
      {\bfseries \large Veronika Solopova$^{1,2}$, Max Upravitelev$^{1,2}$, Daniel Röder$^2$, Iffat Maab$^4$} \\
      {\bfseries \large Junichi Yamagishi$^4$, Sebastian Möller$^{1,2}$, Vera Schmitt$^{1,2}$}}

\address{$^1$Technische Universität Berlin, Quality and Usability Lab, Berlin, Germany \\  $^2$Deutsches Forschungszentrum für Künstliche Intelligenz (DFKI), Berlin, Germany\\
         $^3$Harz University of Applied Sciences, Wernigerode, Germany\\
         $^4$National Institute of Informatics, Tokyo, Japan\\
         \texttt{\{sahitaj, sebastian.moeller, vera.schmitt\}@tu-berlin.de} \\
         \texttt{\{ariana.sahitaj, veronika.solopova\}@tu-berlin.de, daniel.roeder@dfki.de} \\
         \texttt{\{maab, jyamagis\}@nii.ac.jp, u37871@hs-harz.de}}

\abstract{
Fact-checking articles encode rich supporting evidence and reasoning, yet this evidence remains largely inaccessible to automated verification systems due to unstructured presentation. We introduce \texttt{PrimeFacts}, a methodology and resource for extracting fine-grained evidence from full fact-checking articles. We compile 13,106 PolitiFact articles with claims, verdicts, and all referenced sources, and we identify 49,718 in-article hyperlinks as natural anchors to pinpoint key evidence. Our framework leverages large language models (LLMs) to rewrite these anchor sentences into stand-alone, context-independent premises and investigates the extraction of additional implicit evidence. In evaluations on cross-article evidence retrieval and claim verification, the extracted premises substantially improve performance. Decontextualized evidence yields higher retrievability, achieving up to a 30\% relative gain in Mean Reciprocal Rank over verbatim sentences, and using the evidence for verdict prediction raises Macro-F$_1$ by 10-20 points over the baseline. These gains are consistent across different verdict granularities (2-class vs. 5-class) and model architectures. A qualitative analysis indicates that the decontextualized premises remain faithful to the original sources. Our work highlights the promise of reusing fact-checkers’ evidence for automation and provides a large-scale resource of structured evidence from real-world fact-checks. 
\\ \newline  }

\begin{document}

\maketitleabstract

\section{Introduction}
\label{sec:intro}

Professional \textit{fact-checking} has become a crucial process to counter misinformation and disinformation in politics and media. Journalistic fact-checkers investigate a check-worthy claim by gathering supporting and refuting evidence and then issuing a verdict on the claim's veracity \cite{graves2016DecidingWhatsTrue, jiang2020FactoringFactChecksStructured}. The results of this labor-intensive process are published as articles containing the claim, contextual background, evidence discussion, and a final verdict on truthfulness \cite{sahitaj2025AutomatedFactCheckingRealWorld}. Standardization efforts like the ClaimReview schema have encouraged fact-checkers to explicitly mark the claim and verdict in each article, but other critical components, namely the \textit{evidence} that led to the verdict and the \textit{reasoning} connecting evidence to conclusion, are rarely structured or annotated due to the extra workload this would entail \cite{jiang2020FactoringFactChecksStructured}. As a result, rich supporting information in fact-check articles remains locked in unstructured text, limiting its reusability for automated verification systems and obscuring the transparency of how conclusions are reached for a given claim \cite{jiang2020FactoringFactChecksStructured, alhindi2018WhereYourEvidence}.
\begin{figure}
    \centering
    \includegraphics[width=1\linewidth]{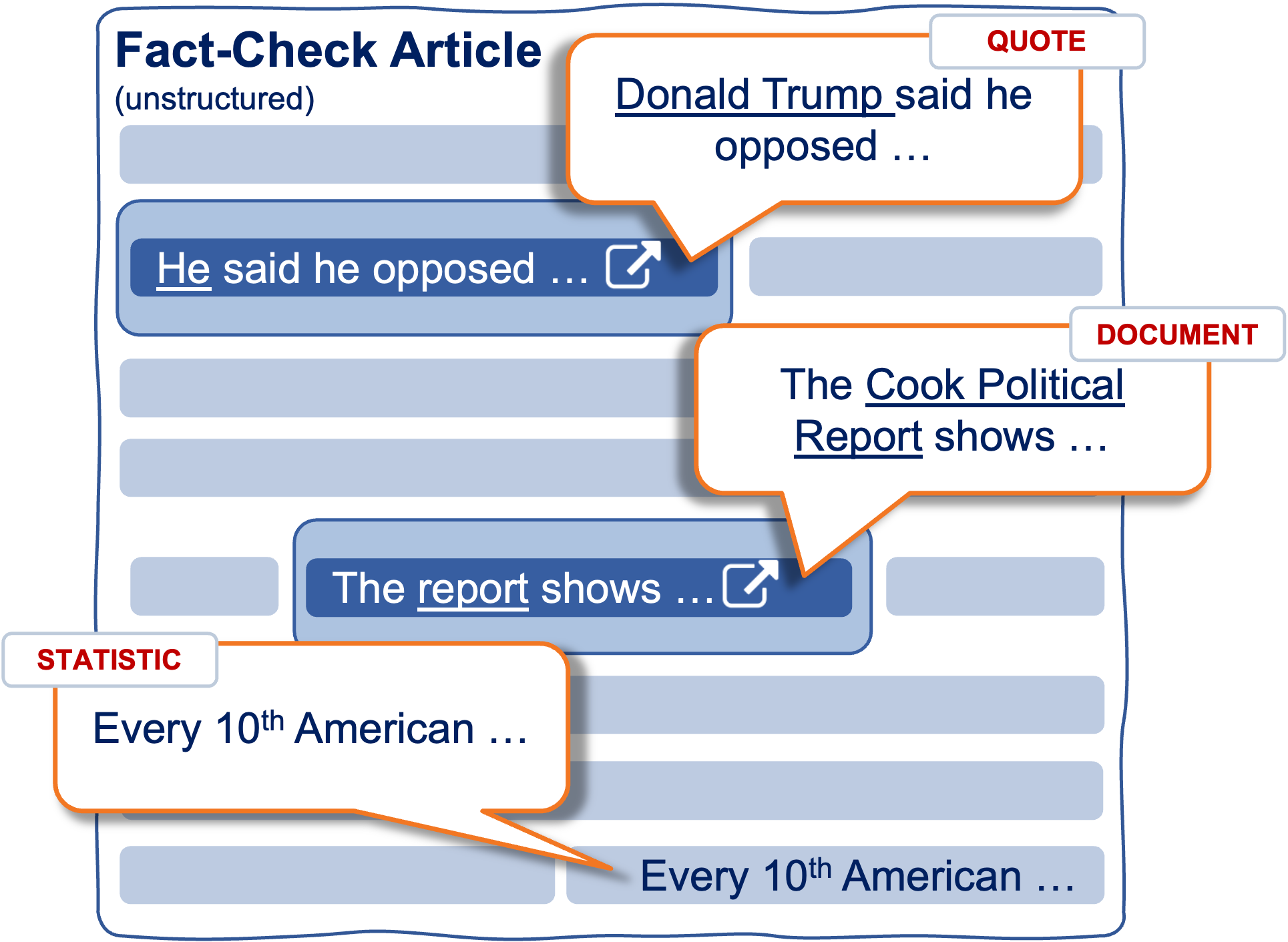}
    \caption{Example fragments from a fact-checking article. (Top and middle) hyperlink-anchored sentence \textit{decontextualized} in \textit{Mode~B}. (Bottom) statistical claim identified by \textit{open extraction} in \textit{Mode~C}.}
    \label{fig:example}
\end{figure}
Similar or rephrased claims tend to reappear across media, causing fact-checkers to spend effort redundantly on already verified information \cite{nakov2021OverviewCLEF2021CheckThat,panchendrarajan2024ClaimDetectionAutomated}. We argue that \emph{evidence reuse} is a valuable extension of claim matching: once supporting material is extracted and normalized, the same evidence can be associated with multiple variants of a claim, enabling future claims to be verified using pre-existing evidence. Moreover, this would enable analyses of aspects such as source diversity and reasoning consistency. For instance, one could examine how premises are assembled to support or refute a target claim, or assess whether the evidence usage reveals informational bias in an article's reasoning \cite{ijcai2025p1274, stab2017parsing, maab-etal-2024-media}. However, manually annotating evidence at scale is impractical. Fact-check articles are often long, densely sourced, and rhetorically complex \cite{humprecht2020HowTheyDebunk,jiang2020factoring}. Exhaustively identifying all relevant evidence sentences and rewriting them as standalone statements would require significant expert effort, leading to annotator fatigue, inconsistency, and prohibitive cost \cite{ostrowski2021MultiHopFactChecking, schlichtkrull2023averitec}. This motivates exploring automated methods to unlock the evidence within fact-check articles.

These articles contain \textit{in-text citations} that serve as natural anchors for evidence to enhance the transparency of the manual verification process (see Figure~\ref{fig:example} for an illustrative example). Specifically, hyperlinks to primary sources such as data, reports and transcripts are extensively embedded within their writing \cite{cazzamatta2025building, humprecht2020HowTheyDebunk}. Therefore, these reference links can be leveraged to automatically pinpoint the key supporting sentences within an article. By extracting and appropriately reformulating those anchor sentences, we aim to make the evidence both \emph{addressable} (tied to a specific source reference) and \emph{portable} (understandable outside the original context) for use in downstream automated retrieval and verification systems. In this work, we investigate the feasibility of automatically extracting evidence from fact-check articles by exploiting their cited sources and language models. We focus on three research questions:

\begin{itemize}
\item \textbf{RQ1 (Addressability):} To what extent can in-article hyperlinks serve as a reliable proxy for identifying the core evidence that supports or refutes a claim?
\item \textbf{RQ2 (Portability):} Does rewriting premises into stand-alone, \emph{decontextualized} statements improve their usefulness for retrieval and automated verification tasks?
\item \textbf{RQ3 (Robustness):} Are the findings consistent across different evaluation settings and task granularities?
\end{itemize}

Our contributions are four-fold: (i) We curate \texttt{PrimeFacts}, a resource of 13{,}106 PolitiFact fact-checks with structured article metadata and 49{,}718 in-article hyperlinks that serve as evidence anchors. (ii) We introduce a three-mode extraction framework: anchored verbatim sentences, LLM-based decontextualization into stand-alone premises, and open extraction of self-contained premises with source attributions. We additionally define a lightweight evidence-type ontology to characterize premise content. (iii) We propose a score for evaluating the faithfulness of decontextualizations, combining forward textual entailment with an asymmetric lexical-overlap penalty, and conduct a targeted human study to assess self-containment and evidence typing. (iv) We evaluate \emph{evidence reuse} on two downstream tasks, cross-article retrieval and zero-shot claim verification, across six instruction-tuned LLMs and two verdict granularities. 

\begin{figure*}
    \centering
    \includegraphics[width=1\linewidth]{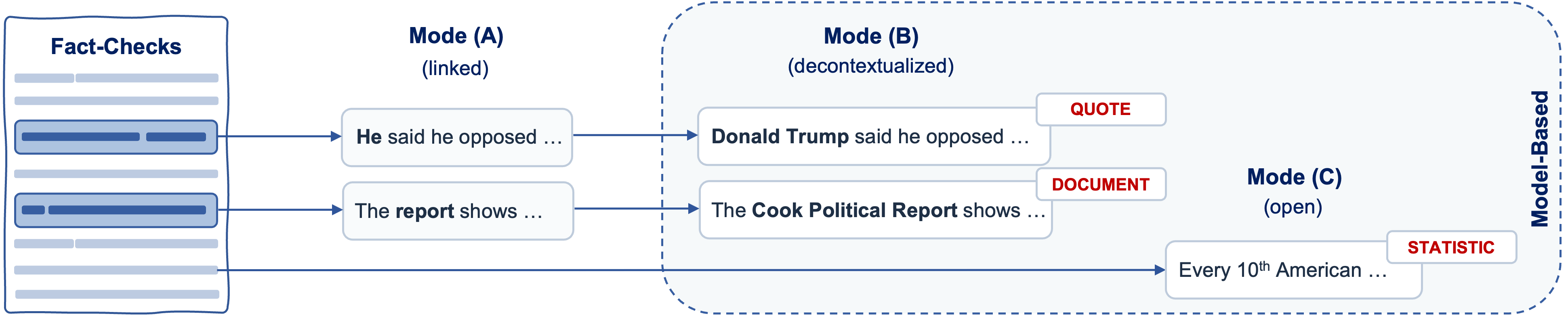}
    \caption{Extraction pipeline for transforming fact-checks into refined, decontextualized evidence.}
    \label{fig:pipeline}
\end{figure*}

\section{Related Work}
\label{sec:relatedwork}

Automated fact-checking (AFC) research aims to verify claims by retrieving evidence and predicting verdicts, often producing a textual explanation \cite{guo2022SurveyAutomatedFactChecking}. A persistent challenge in AFC is the shortage of training data that contains real-world claims paired with gold-standard evidence and detailed reasoning. Many existing datasets use artificially constructed claims or evidence from Wikipedia, e.g. FEVER \cite{thorne2018FEVERLargescaleDataset}, and extensions \cite{schuster2021GetYourVitamin, aly2021FEVEROUSFactExtraction, ma2024EXFEVERDatasetMultihop}, or consider the referenced pages within fact-checking articles as evidence \cite{augenstein2019MultiFCRealWorldMultiDomain, khan2022WatClaimCheckNewDataset}. Other datasets target fact-checking justifications by relying on heuristic extractions \cite{alhindi2018WhereYourEvidence, zeng2024JustiLMFewshotJustification} or are limited in scale due to manual annotation \cite{ostrowski2021MultiHopFactChecking,wang2025WhenAutomatedFactchecking}. These approaches do not isolate or quantify the evidence-bearing sentences that journalists integrate into the article's reasoning. Instead, they ingest entire referenced pages or treat fact-checks as the evidence. Reliance on whole fact-checking articles introduces noise and reduces system effectiveness \cite{samarinas2021improving, xing2024evaluating, deng2025next, sauchuk2022role}, while incomplete or inaccessible sources and link rot further undermine automated evidence retrieval \cite{cazzamatta2025building, warren2025show, kavtaradze2024challenges, zhou2015no, klein2014scholarly}.
\texttt{PrimeFacts} addresses this gap by focusing on evidence units within fact-checking articles that journalists already anchor through in-text hyperlinks and end-of-article references. We promote both precise addressability via these anchored spans and enhanced portability by reformulating them into context-independent premises. Our open extraction approach shares the core insight of \citet{chen2024DenseXRetrieval} that decomposing text into atomic, self-contained propositions improves retrieval, but differs in scope and constraints. \citet{chen2024DenseXRetrieval} define propositions as minimal, self-contained factoids and exhaustively decompose passages so that all propositions together recover the full passage semantics, using a compact model distilled from GPT-4 outputs. In contrast, our approach performs selective, attribution-based extraction from fact-checking articles, bounded by the article's anchor count and targeting key evidence rather than every atomic fact. Recent work on maintaining editable knowledge bases \cite{li2024LanguageModelingEditable} further motivates the extraction of modular, updateable evidence units, a goal that our decontextualized premises directly support.

\section{Data Collection}\label{sec:data}

Each record in \texttt{PrimeFacts} corresponds to a single fact-checking article from the English-language PolitiFact\footnote{\url{https://www.politifact.com}} archive, collected from the chronological and topic-based indexes up to September~2025. We retain only fact-check articles targeting text-based claims, identified via URL and markup patterns, to preserve internal validity and consistent textual structure. The final dataset comprises 13{,}106 fact-check articles authored by 661 individual journalists, each containing at least one in-text hyperlink to an external source. For each article, we store a structured representation linking the canonical PolitiFact URL with its metadata, including crawl timestamp, editorial tags, claim information, author and speaker metadata, and a structured list of cited sources to ensure provenance and reproducibility. The released version of the resource\footnote{\url{https://huggingface.co/datasets/premsa/prime-facts}} contains only derived metadata and annotations. Original fact-check article texts are not redistributed for copyright reasons. The resource is organized into three main components: (i) \textit{Article metadata}, containing the article origin, verdict label, claim statement, and cited materials; (ii) \textit{Entity metadata}, providing unified lookup profiles for authors and speakers to maintain consistent attribution across articles, and (iii) \textit{Annotations}, linking each claim to its extracted anchor statements and aligning them with the corresponding article structure and metadata. All components are cross-referenced through their canonical URLs. A companion statistics file reports aggregate distributions to support stratified sampling and downstream evaluation. Beyond aggregating raw PolitiFact metadata, \texttt{PrimeFacts} contributes several processing steps that constitute a standalone resource: layout-aware text normalization derived from web browser rendering \cite{weichselbraun2021}, sentence-level segmentation with stable letter identifiers, hyperlink extraction and cross-referencing with author-provided source lists, systematic label-leak filtering to prevent verdict contamination, and the decontextualized premise annotations as structured evidence. Together, these transformations turn unstructured fact-check articles into a queryable evidence base that supports retrieval-augmented verification and comparative analysis.

\section{Methodology}\label{sec:methodology}

Before describing the framework, we clarify the key terms used throughout this paper. An \textit{anchor} is an in-article hyperlink that marks a sentence citing an external source. A \textit{source} is the external document or page linked by an anchor (e.g., a government report or dataset). \textit{Evidence} refers broadly to any information supporting or refuting a claim. A \textit{premise} is a decontextualized, self-contained evidence statement derived from the article, suitable for reuse outside its original context. We use these terms consistently hereafter. We propose a framework for extracting fine-grained evidence from fact-check articles and for evaluating its usefulness in downstream verification settings. Figure~\ref{fig:pipeline} gives an overview. The framework includes three evidence modes: hyperlink-anchored sentences, their decontextualized variants, and premises obtained through open extraction. We evaluate the resulting evidence representations in cross-article retrieval and claim verification, and we further assess their faithfulness to the original article content.

\subsection{Evidence Extraction}

Each article is segmented into distinct sentence-like units $u$, each assigned a stable identifier $\iota$. For downstream experiments, we remove sentences that explicitly state the verdict to prevent label leakage, ensuring that models see only evidence, not conclusions \cite{alhindi2018WhereYourEvidence,glockner2022MissingCounterEvidenceRenders}. We then identify anchor sentences that contain at least one hyperlink to an external source. These in-text hyperlinks typically mark factual assertions supported by external evidence \cite{cazzamatta2025building}. The rationale for treating hyperlinked sentences as high-quality evidence markers is grounded in journalistic practice: fact-checkers deliberately embed links to primary sources such as government data, official records, prior reporting, and expert statements to substantiate their analysis and enhance transparency \cite{humprecht2020HowTheyDebunk}. This editorial convention makes hyperlinks natural proxies for evidence-bearing content. We cross-reference each hyperlink with the article's author-provided source list and retain only those pointing to sources that the author has explicitly listed by name in the article's reference section, to ensure we capture only high-quality anchors. We only consider articles with at least one anchor. This leaves us with 13,106 articles and a final set of 49,718 anchors. Using the identified anchors, we derive two sets of evidence, corresponding to Mode~A and its decontextualized variation Mode~B. Mode~C extracts decontextualized premises without anchors. Each mode produces a collection of candidate premises per article with a unique identifier $\iota$ linking it back to the article text for provenance.

\subsubsection{Mode A: Anchored Evidence}

Mode~A uses each anchor sentence verbatim as an evidence unit. This yields a set of factual statements directly grounded in the fact-checker's cited sources. For example, if an article states, "In 2020, the city's homicide rate was the lowest on record" with a hyperlink to a police report, we retain that sentence unchanged as evidence. By design, these anchor-based evidence units are high-precision, but this does not guarantee high recall. However, using them as-is can limit reusability, since anchor sentences may contain pronouns or context-dependent references that are not self-explanatory outside the article context. Mode~A primarily addresses \emph{addressability} (RQ1) and serves as a baseline for our extraction framework.

\subsubsection{Mode B: Decontextualization}

Mode~B aims to improve the \emph{portability} of evidence by rewriting each Mode~A sentence into a stand-alone premise. We employ an LLM to \emph{decontextualize} each anchor sentence, i.e. to make implicit references explicit so that the sentence is understandable out of context \cite{choi2021DecontextualizationMakingSentences}. For example, a Mode~A statement \textit{"He took office in 2019"} might be rewritten as \textit{"Volodymyr Zelenskyy took office as President of Ukraine in May 2019"}, resolving the pronoun and adding context for clarity. Concretely, for each anchor sentence the LLM receives the full letter-segmented article together with the claim statement and the target letter identifier pointing to the anchor sentence. This provides the model with sufficient surrounding context to resolve coreferences and implicit references. We use a zero-shot prompting strategy with structured JSON output guided by a Pydantic schema (see Appendix~\ref{app:prompts}), which constrains the model to return a single decontextualized sentence along with an evidence-type category in one joint generation step. The prompt explicitly instructs the model to preserve the original meaning and factual content while only integrating details necessary for standalone interpretation. This follows best practices from recent work on minimality in decontextualization \cite{gunjal2024MolecularFactsDesiderata}. Each decontextualized premise is output together with its corresponding identifier $\iota$ for traceability to the source unit $u$. Mode~B produces one decontextualized premise for each Mode~A evidence unit. This allows evidence from one fact-check to be more readily understood and reused in verifying other claims, addressing RQ2. Building on prior work investigating evidence operationalization in the fact-checking newsroom by \citet{cazzamatta2025RedefiningObjectivityExploring} and consistent with comparative evidence that fact-checkers routinely add background and context to qualify and make verdicts interpretable \citep{cazzamatta2025DecodingCorrectionStrategies}, we define a novel ontology of evidence types and assign types to each decontextualized premise: \texttt{QUOTE} as attributed statement by a person or organization, \texttt{STATISTIC} as numeric fact from an official dataset or series, \texttt{DOCUMENT} as findings of an authoritative record (e.g., law, ruling, report, prior fact-check), \texttt{CONTEXT} as background attribution or qualification needed to interpret premises, and \texttt{OTHER} if none of the above fits. These labels do not affect the extraction framework, but help characterize what kinds of information the premises convey.

\subsubsection{Mode C: Open Extraction}

Mode~C explores a more open-ended extraction using LLM generation. Instead of relying only on explicit hyperlinks, we prompt an LLM to process the entire fact-checking article and directly output a set of self-contained premises. Prompt details are provided in Appendix~\ref{app:prompts}. The purpose of this approach is to capture any central premises in the article that the journalist may have implied and not explicitly linked. We constrain this process to maintain alignment with the article: the model is instructed to produce at most $n$ premises, where $n$ equals the number of Mode~A anchors in that article, to ensure a fair comparison. For each generated premise, the LLM must cite the supporting unit identifier $\iota$ from which it drew the information. Comparing Mode~C to the anchor-driven modes thus reveals whether important premises were not considered by only looking at hyperlinks. This serves as a stress test for our framework, since any premises found in Mode~C should ideally overlap with or complement Mode~A/B if the anchor-based approach is comprehensive.

\subsection{Evidence Reuse}

\subsubsection{Faithfulness}
\label{sec:fidelity}

Decontextualization rewrites a source sentence into a stand-alone premise that preserves the original factual content while resolving contextual ambiguity \cite{choi2021DecontextualizationMakingSentences}. Prior work in abstractive summarization has shown that Natural Language Inference (NLI) models, which determine if a premise entails, contradicts, or is neutral towards a source, correlate higher with human judgments of factual consistency than standard measures such as \texttt{ROUGE} or \texttt{BERTScore} \cite{maynez2020FaithfulnessFactualityAbstractivea}. We adapt this idea and evaluate the \emph{faithfulness} of a decontextualized premise $p$ to its source sentence $s$ as a NLI problem, using \emph{forward} textual entailment. Forward direction captures the desideratum that a more specific, context-completed rewrite should imply the key facts of the original sentence, while the reverse direction does not need to hold. For example, the premise \textit{"The unemployment rate doubled in 2016, according to the Bureau of Labor Statistics."} correctly entails the more general source statement \textit{"The rate doubled in 2016"}. The reverse is not necessarily true, making this a directional check of information preservation. However, NLI models can be susceptible to lexical-overlap \cite{naik2018StressTestEvaluation}. A premise that simply copies, minimally edits, or truncates $s$ can receive a high entailment score despite failing the decontextualization objective. To address this, we introduce the \textbf{Decontextualization Faithfulness Score (DFS)}, a composite measure that balances factual entailment with an explicit lexical overlap penalty across a dataset $D \subseteq P \times S$, where $P$ and $S$ are the sets of generated premises and source sentences

\[
\begin{aligned}
\mathrm{DFS}(D)
&= \frac{1}{|D|}\sum_{(p,s)\in D}
\mathrm{E}(p, s)\,\bigl(1 - \mathrm{O}(p, s)\bigr), \label{eq:dfs}\\[2mm]
\mathrm{O}(p, s)
&= \frac{\bigl|\operatorname{t}(p)\cap \operatorname{t}(s)\bigr|}
         {\bigl|\operatorname{t}(p)\bigr|},
\end{aligned}
\]

where $\mathrm{E}(p,s)\in[0,1]$ is the probability that $p$ entails $s$, $\mathrm{O}(p,s)\in[0,1]$ is the asymmetric lexical overlap of $p$ covered by $s$, and $\operatorname{t}(\cdot)$ denotes the (multi)set of tokens. The token overlap is normalized by the length of the generated premise $|\operatorname{t}(p)|$, which correctly penalizes premises that are simple excerpts of the source while rewarding the addition of new, contextualizing tokens that improve portability without undermining factual consistency with its source. We compute DFS for both Mode~B and Mode~C across model outputs to evaluate the faithfulness of the decontextualizations.

\subsubsection{Retrievability}

Retrievability operationalizes portability: if a premise is self-contained enough, then, given a claim as a query, standard retrieval methods should locate it more reliably in a corpus of prior fact-checks. To assess how well the resulting premises serve as a reusable knowledge base for claim and evidence matching \cite{panchendrarajan2024ClaimDetectionAutomated}, we simulate the task of retrieving relevant evidence for a new claim by searching a database of verified facts from prior fact-checks. Specifically, for each mode, we compile all premises from the collected articles into a search index. We treat each article's claim statement as a query and attempt to retrieve candidate evidence from the indexed premises of all fact-checks. We compute ranked lists over the mode-specific indexes and score effectiveness with standard information retrieval metrics. By comparing retrieval performance across modes, we aim to quantify the benefit of evidence decontextualization and open extraction on evidence matching. We interpret relative gains from Mode~A$\!\to$B/C as evidence that decontextualization improves cross-article portability.

\subsubsection{Verification Utility}

We investigate whether surfaced premises enable zero-shot claim verification. For each claim, a model receives the premises for that article and the label schema, and must output (i) a verdict from the allowed set and (ii) a brief justification that cites used premises via identifiers $\iota$ \cite{guo2022SurveyAutomatedFactChecking}. Cited IDs make decisions traceable and let us quantify evidence use \cite{jolly2022GeneratingFluentFact}. Alongside verdict accuracy, we report \emph{citation coverage} as the fraction of presented premises that are cited: with $S_{\text{given}}$ the shown premise IDs and $S_{\text{cited}}\subseteq S_{\text{given}}$ those mentioned, $C = |S_{\text{cited}}|/|S_{\text{given}}|$. Coverage contrasts evidence modes as more informative premises should improve task performance while citing fewer items. Accordingly, we treat coverage as diagnostic rather than a target, since correctness is our primary objective.

\section{Experiments}

\subsection{Setup}

We evaluate our approach on the collected and processed corpus of 13,106 PolitiFact fact-check articles (Section~\ref{sec:data}). Each fact-check instance provides a query claim and its extracted evidence set per mode. Because our method does not require model fine-tuning, we do not partition the data into training splits. Instead, all evidence extraction and verification experiments are conducted in a zero-shot inference setting with LLMs, respectively. We compare six publicly available state-of-the-art instruction-tuned LLMs of varying scale and architecture: Qwen3 at 8B, 14B, and 32B dense and 235B mixture of experts (MoE) with 22B active during inference \cite{yang2025qwen3}, Llama 3.3 at 70B dense \cite{llama3}, and Llama 4 Scout MoE with 109B total and 17B active \cite{llama4}.

\subsection{Automatic Evaluation}

We operationalize the three research questions through complementary evaluations: retrieval performance measures portability (RQ2) by testing whether decontextualized premises are more discoverable across articles; verification performance measures whether portable evidence also improves downstream task accuracy; and faithfulness analysis ensures that LLM-based rewriting preserves factual content. By evaluating across six LLMs of varying scale and architecture and two verdict granularities (binary and five-class), we assess robustness (RQ3). We acknowledge that model-size differences confound model-specific and knowledge-driven effects and discuss this in Limitations.

\subsubsection{Retrieval Performance}

Each claim statement is used as a query to retrieve evidence sentences from an index of all extracted premises (Section~\ref{sec:methodology}), simulating cross-article evidence reuse. We use BM25 to construct an efficient retrieval index \cite{lu2024BM25SOrdersMagnitude}. We measure standard ranking metrics: Mean Reciprocal Rank at 10 (MRR@10), normalized Discounted Cumulative Gain (nDCG) at 3 and 10, and Recall (R) at 1, 3, 10, treating the premises from the claim's own fact-check as the relevant gold truth set. Higher MRR and nDCG indicate that the relevant evidence is ranked near the top, while higher Recall@$k$ indicates more of the gold premises are retrieved within the top $k$ results.

\begin{table}[t]
  \centering
  \small
  \tabcolsep=0.10cm
  \begin{tabular}{lcccccc}
    \toprule
     & \multicolumn{1}{c}{\textbf{MRR}} & \multicolumn{2}{c}{\textbf{nDCG}} & \multicolumn{3}{c}{\textbf{Recall}} \\
     \cmidrule(lr){2-2} \cmidrule(lr){3-4} \cmidrule(lr){5-7}
     \textbf{Model/Mode} & 10 & 3 & 10 & 1 & 3 & 10 \\
    \midrule
     \textbf{Baseline} A & 0.43 & 0.26 & 0.23 & 0.10 & 0.15 & 0.21 \\
    \midrule
     \textbf{Qwen3-8B} & & & & & & \\
     \hspace{2mm}{B (decontext.)} & 0.47 & 0.30 & 0.27 & 0.11 & 0.18 & 0.25 \\
     \hspace{2mm}{C (open)} & 0.88 & 0.67 & 0.62 & 0.30 & 0.46 & 0.57 \\
    \midrule
     \textbf{Qwen3-14B} & & & & & & \\
     \hspace{2mm}{B (decontext.)} & 0.46 & 0.28 & 0.25 & 0.11 & 0.17 & 0.23 \\
     \hspace{2mm}{C (open)} & 0.81 & 0.59 & 0.54 & 0.27 & 0.40 & 0.50 \\
    \midrule
     \textbf{Qwen3-32B} & & & & & & \\
     \hspace{2mm}{B (decontext.)} & 0.50 & 0.32 & 0.29 & 0.12 & 0.20 & 0.27 \\
     \hspace{2mm}{C (open)} & 0.78 & 0.56 & 0.51 & 0.24 & 0.38 & 0.48 \\
    \midrule
     \textbf{Qwen3-235B} & & & & & & \\
     \hspace{2mm}{B (decontext.)} & 0.58 & 0.40 & 0.37 & 0.15 & 0.25 & 0.35 \\
     \hspace{2mm}{C (open)} & 0.81 & 0.62 & 0.57 & 0.26 & 0.41 & 0.54 \\
    \midrule
     \textbf{Llama-4-Scout} & & & & & & \\
     \hspace{2mm}{B (decontext.)} & 0.49 & 0.31 & 0.28 & 0.12 & 0.19 & 0.27 \\
     \hspace{2mm}{C (open)} & 0.80 & 0.59 & 0.54 & 0.26 & 0.40 & 0.50 \\
    \midrule
     \textbf{Llama-3.3-70B} & & & & & & \\
     \hspace{2mm}{B (decontext.)} & 0.59 & 0.41 & 0.38 & 0.15 & 0.25 & 0.36 \\
     \hspace{2mm}{C (open)} & 0.84 & 0.63 & 0.57 & 0.28 & 0.43 & 0.53 \\
    \midrule
  \end{tabular}
  \caption{Retrieval results across models.}
  \label{tab:retrieval_abc_models_compact}
\end{table}

We first examine the effectiveness of evidence retrieval across different extraction modes. Table~\ref{tab:retrieval_abc_models_compact} compares decontextualized premises from Mode~B against the baseline using verbatim sentences from Mode~A. Decontextualization yields substantial gains in all metrics for every model. For instance, Llama-3.3-70B achieves an MRR@10 of 0.59 with Mode~B, compared to 0.43 for the baseline, which is a 37\% improvement. Recall@10 improves from 0.21 to 0.36, meaning the self-contained premises allow 71\% more of the relevant evidence to be retrieved in the top-10 results. We observe consistent improvements at rank-3 as well (nDCG@3 from 0.26 to 0.41). These results confirm that making evidence sentences context-independent greatly increases their \textit{portability} and discoverability by lexical-matching methods, addressing RQ2.
Table~\ref{tab:retrieval_abc_models_compact} also shows retrieval results for Mode~C with LLM-generated premises without anchor cues. Mode~C outperforms the baseline by a wide margin. Across models, MRR@10 ranges from 0.78 to 0.88, indicating that a large share of queries have a relevant premise at rank 1. For example, Qwen3-8B and Qwen3-14B reach MRR@10 scores of 0.88 and 0.81, respectively. Recall@10 more than doubles relative to the baseline, reaching up to 0.57, with all models at 0.48 or higher, meaning that Mode~C premises capture a larger portion of the self-selected evidence per claim. The retrieval scores for Mode~C suggest that the LLMs extract evidence that is more directly aligned with the claim than anchored sentences from Mode~A. This highlights a potential trade-off. While Mode~C yields high recall and may surface additional relevant premises beyond explicit hyperlink anchors, it may also benefit from more direct lexical overlap with the claim wording. We return to this point in Section~\ref{sec:discussion}. Overall, the trend from Mode~A $\to$ B $\to$ C is one of strictly increasing retrieval effectiveness, demonstrating the benefit of evidence decontextualization and open extraction.

\subsubsection{Verification Performance}
\label{sec:verification-performance}

We evaluate label prediction using Macro-F$_1$ for both a binary setting, omitting \texttt{half-true} in the binary collapse, and a fine-grained five-class setting. Macro-F$_1$ is appropriate due to class imbalance, see Table~\ref{tab:all_class_dist}, and the need to reward balanced performance across all verdict categories. We also quantify the evidence usage in each model's explanation by computing the citation coverage. Table~\ref{tab:results_abcd_models_three_five} reports Macro-F$_1$ scores for each model and evidence mode, for both the binary and five-class verdict prediction tasks. Several clear patterns emerge. First, providing any evidence (Mode~A) dramatically improves performance over the majority-class baseline which achieves Macro-F$_1$ of 0.39 for binary and 0.10 for five-class. Even the raw anchor sentences enable Macro-F$_1$ in the 0.57-0.68 range (binary) depending on the model, confirming that journalist-provided reference sentences capture relevant factual information needed to judge veracity. This supports RQ1.

Second, decontextualizing the evidence consistently boosts accuracy over Mode~A. All models see an absolute Macro-F$_1$ gain of 4-7 points in the binary setting and up to 8 points in the five-class setting when using self-contained premises from Mode~B. For example, Qwen3-32B improves from 0.59 to 0.66 (binary) and 0.27 to 0.28 (five-class). The largest jump is for Llama-3.3-70B, rising to 0.74 (binary) and 0.35 (five-class) in Mode~B, a relative improvement of $\sim$8 and $\sim$27 percent, respectively. This indicates that evidence portability (RQ2) is not only beneficial for retrieval, but also aids the model in understanding and applying the evidence to the claim. By reducing ambiguity through operations such as resolving pronouns or making implicit context explicit, decontextualized premises make it easier for the verification model to connect facts to the claim. 

\begin{table}[t]
    \centering
    \small
    \tabcolsep=0.15cm
    \begin{tabular}{llcccc} 
        \toprule
        \multirow{1}{*}{\textbf{\textsc{}}} & \multicolumn{2}{c}{\textbf{5-class}} & \multicolumn{2}{c}{\textbf{2-class}} \\ 
        \cmidrule(lr){2-3} \cmidrule(lr){4-5}
        \textbf{Labels}  & \textbf{Count} & \textbf{\%}  & \textbf{Count} & \textbf{\%} \\
        \midrule
        true          & 1{,}513 & 11.5\% & \textemdash    & \textemdash    \\
        mostly-true   & 2{,}283 & 17.4\% & 3{,}794 & 35.6\% \\
        half-true     & 2{,}443 & 18.6\% & \textemdash   & \textemdash    \\ 
        mostly-false  & 2{,}425 & 18.5\% & 6{,}860 & 64.4\% \\ 
        false         & 4{,}442 & 33.9\% & \textemdash    & \textemdash    \\
        \bottomrule
    \end{tabular}
    \caption{Distribution of data for five- and two-class settings.}
    \label{tab:all_class_dist}
\end{table}

Third, Mode~C yields the highest verification performance across the board. Qwen3-235B and Llama-3.3-70B both reach binary Macro-F$_1$ scores of 0.81, and Llama-3.3-70B achieves the top five-class score of 0.42. Relative to Mode~B, this corresponds to an additional gain of 12 points for Qwen3-235B in the binary setting and 7 points for Llama-3.3-70B in the five-class setting. Notably, the improvements from Mode~B to Mode~C are smaller than from Mode~A to Mode~B, suggesting diminishing returns and possible overlap between anchor-based and open-extracted evidence. On further investigation, we find that, on average, about 25\% of the source references surfaced in Mode~C overlap with anchor-based evidence from Mode~A. The trend holds across all model sizes and for both binary and fine-grained tasks, supporting the robustness hypothesis in RQ3. We also observe that larger models tend to perform better overall. For instance, Llama-3.3-70B outperforms the smaller Qwen3 models in each mode, which is expected given its greater parametric knowledge.

\begin{table}[t]
  \centering
  \small
  \tabcolsep=0.10cm
  \begin{tabular}{lcccc}
    \toprule
     & \multicolumn{2}{c}{\textbf{Two Class}} & \multicolumn{2}{c}{\textbf{Five Class}} \\
     \cmidrule(lr){2-3} \cmidrule(lr){4-5}
     \textbf{Model/Mode}    & {\scriptsize \textbf{Macro-F1}} & {\scriptsize \textbf{Coverage}} & {\scriptsize \textbf{Macro-F1}} & {\scriptsize \textbf{Coverage}} \\
    \midrule
       \textbf{ Baseline }          &   0.39        &    \textemdash    &  0.10        &     \textemdash   \\
    \midrule
   \textbf{ Qwen3-8B }          &           &          &          &         \\
    \hspace{2mm}{ A (linked)}      & 0.58 & 0.82   & 0.21    & 0.75 \\
    \hspace{2mm}{ B (decontext.)}  & 0.63 & 0.64   & 0.22    & 0.63 \\
    \hspace{2mm}{ C (open)}        & 0.73 & 0.72   & 0.25    & 0.71 \\
    \midrule
    \textbf{Qwen3-14B  }        &           &          &          &        \\
    \hspace{2mm}{ A (linked)}      & 0.57 & 0.78   & 0.23    & 0.74 \\
    \hspace{2mm}{ B (decontext.)}  & 0.61 & 0.65   & 0.27    & 0.63 \\
    \hspace{2mm}{ C (open)}        & 0.72 & 0.77   & 0.34    & 0.74 \\
    \midrule
    \textbf{Qwen3-32B }          &          &         &          &         \\
    \hspace{2mm}{ A (linked)}      & 0.59 & 0.71   & 0.27    & 0.71 \\
    \hspace{2mm}{ B (decontext.)}  & 0.66 & 0.64   & 0.28    & 0.64 \\
    \hspace{2mm}{ C (open)}        & 0.76 & 0.75   & 0.33    & 0.75 \\
    \midrule
    \textbf{Llama-4-Scout} &         &          &          &       \\
    \hspace{2mm}{ A (linked)}      & 0.65 & 0.74   & 0.27    & 0.74 \\
    \hspace{2mm}{ B (decontext.)}  & 0.70 & 0.65   & 0.27    & 0.65 \\
    \hspace{2mm}{ C (open)}        & 0.76 & 0.79   & 0.34    & 0.78 \\
    \midrule
    \textbf{Llama-3.3-70B  }       &        &          &          &      \\
    \hspace{2mm}{ A (linked)}      & 0.68 & 0.93   & 0.28    & 0.93 \\
    \hspace{2mm}{ B (decontext.)}  & 0.74 & 0.78   & 0.35    & 0.77 \\
    \hspace{2mm}{ C (open)}        & \textbf{0.81} & 0.86   & \textbf{0.42}    & 0.85 \\
    \midrule
    \textbf{Qwen3-235B}         &          &         &          &        \\
    \hspace{2mm}{ A (linked)}      & 0.62 & 0.82   & 0.26    & 0.83 \\
    \hspace{2mm}{ B (decontext.)}  & 0.69 & 0.70   & 0.30    & 0.46 \\
    \hspace{2mm}{ C (open)}        & \textbf{0.81} & 0.81   & 0.39    & 0.79 \\
    \midrule
  \end{tabular}
  \caption{Results across extraction modes (A-C) and models. Bold values indicate best Macro-F$_1$ per setting.}
  \label{tab:results_abcd_models_three_five}
\end{table}

\subsubsection{Evidence Faithfulness}
\label{sec:evidence-fidelity}

As defined in Section~\ref{sec:fidelity}, we quantify faithfulness with the \emph{Decontextualization Faithfulness Score} (DFS), which combines forward textual entailment $E$ with an explicit penalty for lexical copy-overlap. In all experiments, $E$ is estimated by a standard DeBERTa-Large cross-encoder fine-tuned on SNLI \cite{bowman2015LargeAnnotatedCorpus} and MNLI \cite{williams2018BroadCoverageChallengeCorpus}. Table~\ref{tab:dfs_bc} reports mean forward entailment ($E$) and DFS for \emph{Mode~B} between anchored source sentences and their decontextualized premises and \emph{Mode~C} between the referenced source sentences and open-extracted premises. \textit{Caveat:} DFS can underestimate quality when the original source sentence is already self-contained and well decontextualized, because high lexical overlap $O$ depresses the score even if $E$ is strong. Across models, three patterns emerge. First, in Mode~B, smaller and mid-sized models achieve very high entailment with low DFS, indicating mostly minimal edits (e.g., Qwen3-14B has $E{=}0.91$ but DFS${=}0.03$, and Qwen3-8B $E{=}0.81$ with DFS${=}0.06$), whereas Qwen3-235B and Llama-3.3-70B strike a better balance with higher DFS (0.19 and 0.21) at moderate $E$ (0.76 and 0.67), suggesting more substantive, portable rewrites rather than near-copies. Second, in Mode~C, as expected for more abstractive generation, $E$ decreases across models while DFS rises for some configurations, indicating non-trivial reformulations. Qwen3-235B attains the strongest DFS in Mode~C (0.16), followed by Qwen3-8B (0.11) and Llama-3.3-70B (0.09), reflecting premises that are less verbatim yet still sufficiently supported to aid downstream use. Notably, Mode~B outputs appear more faithful to their references than Mode~C, plausibly because the two-step, anchor-driven pipeline with the anchor selection followed by constrained decontextualization biases rewrites toward the cited source, whereas open extraction has more freedom to abstract and synthesize. Third, forward entailment alone tends to overestimate trivial copy-edits, while DFS differentiates portable decontextualizations from near-verbatim text. Models with higher DFS in Mode~B (Qwen3-235B, Llama-3.3-70B) also yield strong retrieval and verification results (Table~\ref{tab:retrieval_abc_models_compact}, Table~\ref{tab:results_abcd_models_three_five}), aligning faithfulness with utility, and although Mode~C is more abstractive (lower $E$), its DFS indicates that many generated premises remain useful as complementary element to anchor-driven evidence.

\begin{table}[b]
  \centering
  \setlength{\tabcolsep}{0.12cm}
  \small
  \begin{tabular}{lcccc}
    \toprule
    & \multicolumn{2}{c}{\textbf{Mode B}} & \multicolumn{2}{c}{\textbf{Mode C}} \\
    \cmidrule(lr){2-3}\cmidrule(lr){4-5}
    \textbf{Model} & $E$ & DFS & $E$ & DFS \\
    \midrule
    Qwen3-8B      & 0.81 & 0.06 & 0.50 & 0.11 \\
    Qwen3-14B     & \textbf{0.91} & 0.03 & \textbf{0.65} & 0.06 \\
    Qwen3-32B     & 0.86 & 0.09 & 0.60 & 0.07 \\
    Qwen3-235B    & 0.76 & 0.19 & 0.50 & \textbf{0.16} \\
    Llama-4-Scout & 0.69 & 0.07 & 0.42 & 0.05 \\
    Llama-3.3-70B & 0.67 & \textbf{0.21} & 0.39 & 0.09 \\
    \bottomrule
  \end{tabular}
  \caption{Results for Mean Forward Entailment and DFS for Mode~B and Mode~C. Bold values indicate best performance per metric.}
  \label{tab:dfs_bc}
\end{table} 

\subsection{Human Study}
\label{sec:human-study}
To complement the automatic metrics, we conducted a manual annotation study of the evidence to evaluate extraction utility. We randomly sampled 100 premises each from Mode~B and Mode~C outputs. The sample covered premises extracted by the strongest overall model, Qwen3-235B, to evaluate best-case outputs. Each article contributed at most one sampled item to diversify topics. Two annotators with a background in fact-checking independently labeled all items, using annotation guidelines, and were tasked with assessing (a) whether the statement is self-contained and interpretable without surrounding context, and (b) the evidence type assigned to the statement: \textit{Document}, \textit{Statistic}, \textit{Quote}, or \textit{Context}. Question (a) was rated on an ordinal scale from incomplete (1) to complete (3). For Mode~B (a), the results show an observed agreement rate of 0.87 and a Krippendorff's alpha of 0.255 due to both annotators agreeing on a majority of cases to be self-contained (3). For Mode~B (b), we measure an observed agreement of 0.58 and a Krippendorff's alpha of 0.441 due to significant disagreements on the \texttt{CONTEXT} label. For Mode~C (a), the results show an observed agreement rate of 0.835 and a Krippendorff's alpha of 0.474. For Mode~C (b), we measure an observed agreement of 0.67 and a Krippendorff's alpha of 0.561. After resolving evidence-type annotation disagreements through discussion, with the final label restricted to one of the two original annotator choices, Qwen3-235B achieves a Macro-F$_1$ of 0.859 for Mode~B and 0.857 for Mode~C. Furthermore, we did not identify label leakage or factual inconsistencies in either Mode~B or Mode~C within the annotated samples.

\section{Discussion}
\label{sec:discussion}

Our results show that fact-checking articles contain reusable evidence that can be systematically unlocked. In-text hyperlinks provide a strong and scalable signal for locating evidence-bearing statements, and decontextualizing these statements into stand-alone premises consistently improves both retrieval and verification. This suggests that fact-checkers’ sourcing practices can be repurposed to build structured evidence resources for automated fact-checking. In this sense, \texttt{PrimeFacts} can be interpreted as an intermediate layer between document retrieval and claim verification. Instead of reasoning directly over long source documents, verification models operate on compact, decontextualized premises that explicitly encode the factual content of the evidence.

The comparison between Modes~B and~C reveals a clear trade-off. Mode~B provides grounded, source-linked premises that remain close to the journalist’s explicitly cited evidence, while Mode~C often surfaces additional relevant premises beyond hyperlink anchors and achieves the strongest downstream performance. At the same time, open extraction can introduce redundancy or produce statements with high lexical overlap to the claim itself. On average, about 25\% of the source references surfaced in Mode~C overlap with anchor-based evidence from Mode~A, indicating partial but non-trivial complementarity rather than simple duplication. In practice, this suggests a hybrid strategy: use Mode~B as a faithful foundation and supplement it with non-redundant Mode~C premises to improve coverage.

We also observe that decontextualized premises often support stronger predictions while requiring fewer cited items in the generated justifications. This suggests that self-contained premises may be individually more informative for decision-making, although citation coverage should be interpreted cautiously because it also reflects model selection behavior. One explanation is that decontextualization removes discourse dependencies that would otherwise require models to reconstruct context from surrounding text. By converting evidence into self-contained premises, the reasoning task becomes closer to structured factual inference than long-context interpretation. Faithfulness analyses further indicate that stronger models tend to produce supported rewrites rather than verbatim copies, which aligns with their improved downstream verification utility.

These trends were consistent across model families and across both binary and five-class verdict settings, indicating that the gains stem from the evidence representation rather than from a single model architecture. Overall, \texttt{PrimeFacts} shows that evidence extracted from professional fact-checks can support retrieval-augmented and semi-automated verification workflows, although human oversight remains important when generated premises are used in high-stakes settings.

\section{Conclusion}

We presented \texttt{PrimeFacts}, a methodology and resource for transforming full-length fact-checking articles into a reusable evidence resource, and demonstrated its value for automated misinformation detection. Our framework leverages fact-checkers' own sourcing practices by using hyperlink-anchored evidence, decontextualizing these statements into stand-alone premises, and investigating whether similar evidence can also be extracted without relying on anchors. This yields structured evidence representations that are suitable for downstream retrieval and verification. Our findings support the core assumptions of the paper. First, in-article hyperlinks provide a strong and scalable signal for identifying evidence-bearing content. Second, rewriting anchored evidence into decontextualized premises substantially improves both cross-article retrieval and verdict prediction. Third, these improvements remain consistent across different verdict granularities and model architectures. Together, these results show that evidence extracted from professional fact-checks can serve as an effective intermediate representation between long-form journalistic articles and automated claim verification.

By introducing the \texttt{PrimeFacts} resource and extraction methodology, we aim to support future research on retrieval-augmented fact-checking, evidence reuse, and transparent decision support. More broadly, our work suggests that professional fact-checks are not only useful as final verdicts, but also as rich repositories of structured evidence that can support more transparent, reusable, and effective verification systems.

\section*{Ethical Considerations}\label{sec:ethics}

Our target is to develop \texttt{PrimeFacts} as a structured knowledge base for intelligent decision-support systems in fact-checking and related applications. While it enables automated evidence retrieval and verdict prediction, these functions are designed to assist rather than replace human judgment, particularly in high-stakes or politically sensitive contexts. Collaborative human oversight remains essential to interpret nuance, context, and evolving facts. The evidence extracted in \texttt{PrimeFacts} reflects how fact-checkers present and justify information within their articles. Both the selection and presentation of evidence may encode subtle biases from the authors, such as framing, emphasis, or omission of counterpoints, which our extraction pipeline may in turn reproduce. Similarly, while fact-checking organizations are reputable and adhere to editorial standards, their verdicts and accompanying justifications are not free from subjective interpretation and editorial policies. These judgments can be influenced by institutional perspectives, available sources, or political context. Analyses or systems built on this resource should therefore explicitly account for such potential biases. The dataset is derived from copyrighted fact-checking articles. We publicly release only derived metadata and annotations. Original fact-check article texts are not redistributed for copyright reasons.

\section*{Limitations}
Our work has some limitations that suggest avenues for future work. One primary limitation is the reliance on explicit in-text citations (hyperlinks) to identify evidence. While PolitiFact articles are richly linked, some fact-checks or segments rely on implicit evidence or general knowledge that is not captured by a specific hyperlink. Our pipeline would miss such uncited yet important premises. In domains or languages where fact-checkers provide fewer references, a hybrid extraction strategy, combining the anchor-based method with additional open extraction, may be necessary to achieve high recall. Another limitation lies in the scope of the extracted evidence. We isolate individual supporting sentences but do not explicitly capture the logical structure or multi-hop reasoning that a fact-checker might apply across an article. For example, an article might piece together two separate facts to reach a conclusion, but our current method would list these facts separately without representing their inferential connection. This could limit the usefulness of the evidence in tasks requiring joint reasoning. Future extensions should link premises into argumentative chains and label their roles, enabling a closer mirroring of human reasoning steps. The use of large language models for evidence rewriting and generation introduces additional considerations. Although we took measures to preserve faithfulness, such as constrained prompting and post-hoc entailment checks, LLMs can occasionally produce subtly altered or extraneous details. In our manual evaluation we did not observe major factual errors, but there remains a risk of hallucination, especially as we push the models to be more abstractive across long contexts. Users of the \texttt{PrimeFacts} framework should treat decontextualized premises as suggestions to be compared against the original anchor or reference statements. Furthermore, our evaluation is conducted exclusively on PolitiFact, a single English-language fact-checking platform with a particular editorial style and sourcing convention. While Mode~C is platform-agnostic by design and Mode~B requires only that articles contain hyperlinks, we have not yet validated our pipeline on other platforms (e.g., Full Fact, Snopes) or languages. Cross-platform and multilingual evaluation is planned as future work. We also note that our operationalization of portability through retrieval performance and robustness through multi-model, multi-granularity evaluation may not capture all aspects of these concepts. Model-size differences in our LLM comparisons introduce a confound, since larger models have both more parametric knowledge and better instruction-following ability; disentangling these factors is an open question. Finally, we have not tested robustness to adversarial noise or contradictory evidence in the premise set, which would be a valuable stress test for future work.

\section*{Acknowledgments}
This work is funded by the German Federal Ministry for Research, Technology and Aeronautics (BMFTR) in the scope of the projects \textit{news-polygraph} (03RU2U151C) and \textit{FAR-REASONING} (16IS23068). This work is supported by JST CREST Grants (JPMJCR20D3), Japan.

\section*{References}\label{sec:reference}

\bibliographystyle{lrec2026-natbib}
\bibliography{references_PS,references_AS,references_JK}


\appendix
\section{Appendix}
\label{app:prompts}


\subsection{Mode~B: Decontextualization Prompt}
\label{app:prompt-modeb}

For each anchor sentence, the LLM receives a system prompt followed by a user message. The system prompt instructs the model to produce a single decontextualized sentence with an evidence-type category via structured JSON output:

\begin{quote}
\small
\textbf{System:} ``You are a careful scientific editor. Produce ONE decontextualized sentence that stands alone, explicitly preserving or adding entities, numbers, dates that make the sentence clear even when read outside of the article. Assign a category label using exactly one of: QUOTE, STATISTIC, DOCUMENT, CONTEXT, OTHER. [\ldots category guide with definitions and examples\ldots] Return JSON only.''

\medskip
\textbf{User:} ``Claim: \{claim\} \textbackslash n Article (labeled): \{segmented\_article\} \textbackslash n Target letter: \{letter\} \textbackslash n Target sentence: \{target\_sentence\} \textbackslash n Return JSON only.''
\end{quote}

The JSON schema constrains the output to three fields: \texttt{letter} (the segment identifier), \texttt{decontextualized} (the rewritten sentence), and \texttt{category} (one of the five evidence types).

\subsection{Mode~C: Open Extraction Prompt}
\label{app:prompt-modec}

Mode~C receives the full article and extracts multiple premises at once. The system prompt specifies a bounded range of premises and uses the same category guide:

\begin{quote}
\small
\textbf{System:} ``You are a careful scientific editor. Extract \{min\}--\{max\} non-redundant key premises from the labeled article. For each premise, provide: (a) exactly ONE letter anchor from the article that supports it; (b) ONE decontextualized sentence that stands alone; and (c) a category using exactly one of: QUOTE, STATISTIC, DOCUMENT, CONTEXT, OTHER. [\ldots category guide\ldots] Output JSON only.''

\medskip
\textbf{User:} ``Claim: \{claim\} \textbackslash n Article (labeled): \{segmented\_article\} \textbackslash n Return JSON only.''
\end{quote}

The output schema constrains the response to a list of premise objects, each with \texttt{letter}, \texttt{decontextualized}, and \texttt{category} fields. The maximum list length is set to the number of Mode~A anchors for that article.

\end{document}